\documentclass[letterpaper, 10 pt, conference]{ieeeconf}  
\IEEEoverridecommandlockouts    
\usepackage{cite}
\usepackage{hyperref}
\usepackage{graphicx}
\usepackage{makecell}
\usepackage{multirow}
\usepackage{colortbl}
\usepackage{dcolumn}
\usepackage{amsmath}
\usepackage{booktabs,array,stfloats}

\begin{document}

\title{
A Low-Cost Teleoperable Surgical Robot with a Macro-Micro Structure and a Continuum Tip for Open-Source Research
}

\author{Lachlan~Scott, Tangyou~Liu,~and~Liao~Wu,~\IEEEmembership{Member,~IEEE}
\thanks{This work was supported in part by the Australian Research Council under Grant DP210100879, in part by Heart Foundation under Vanguard Grant 106988, and in part by UNSW Engineering under GROW Grant PS69063.}
\thanks{L. Scott, T. Liu, and L. Wu are with the School of Mechanical and Manufacturing Engineering, University of New South Wales, Sydney, Australia. {\tt\small dr.liao.wu@ieee.org}}
}



\maketitle
\thispagestyle{empty}
\pagestyle{empty}

\begin{abstract}

Surgical robotic systems equipped with micro-scale, high-dexterity manipulators have shown promising results in minimally invasive surgery (MIS).
One barrier to the widespread adoption of such systems is the prohibitive cost of research and development efforts using current state-of-the-art equipment.
To address this challenge, this paper proposes a low-cost and modifiable tendon-driven continuum manipulator for MIS applications.
The device is capable of being teleoperated in conjunction with a macro-scale six-axis robotic arm using a haptic stylus.
Its control software incorporates and extends freely available and open-source software packages.
For verification, we perform teleoperation trials on the proposed continuum manipulator using an electromagnetic tracker. We then integrate the manipulator with a UR5e robotic arm.
A series of simulated tumour biopsies were conducted using the integrated robotic system and an anatomical model (phantom), validating its potential efficacy in MIS applications. The complete source code, CAD files for all additively manufactured components, a parts list for the manipulator, and a demonstration video of the proposed system are made available in this work.

\end{abstract}

\section{Introduction}
Minimally invasive surgery (MIS) refers to surgical practice with the intention of reducing the impact of procedures on patient safety and comfort by avoiding unnecessary damage to patient anatomy. Such procedures are carried out by guiding a tool, or a multi-tool assembly, through a narrow opening to operate on a target area. The opening may be created via an incision, or access may be possible through a natural orifice or lumen \cite{ClarkOpenSource}.

The use of robotic manipulators in surgical procedures presents a myriad of benefits to both patients and practitioners.
In particular, robotic manipulators can achieve extremely high degrees of accuracy in surgical instrument placement, and dexterous manipulation of tissue is resilient to operational hazards such as ionizing radiation \cite{surgicalrobotsandcomputer}.
Additionally, robotic surgical systems can reduce the physical and mental burden experienced by physicians performing precise, coordinated movements for extended periods of time during operation \cite{surgicalrobotsandcomputer}.

The term ``macro-micro robotic system" refers to a design in which a macro-scale robotic manipulator with a relatively
large workspace and high payload capacity operates in conjunction with a micro-scale manipulator functioning as an
end-effector. The micro manipulator is able to perform precision tasks such as those required in MIS procedures while
avoiding damage to operation sites \cite{surgicalrobotsandcomputer, yangmacromicro, continuumrobotssurvey}.
Macro-micro systems effectively aim to address the shortcomings of their component
manipulators while offering a number of the advantages possessed by each \cite{yangmacromicro,yousefmacromicro}.

Current commercial solutions for teleoperated robotic surgery can cost between \$500,000 to \$1.5 million.\cite{Telesurgery}
While previous work has found success in integrating surgical robotic manipulators with robotic arms, these efforts have generally been applied to such high-cost, state-of-the-art platforms, some of which are controlled by closed-source software\cite{LiuHaptics,OptimalVision}. Such systems achieve the accuracy and reliability required of surgical tools. However, research in the field, particularly in developing countries, is limited by such prohibitive costs and restricted access \cite{ClarkOpenSource}\cite{ChenOpenSource}. Open-source, research-grade solutions such as the Raven II platform \cite{raven} exist, however a complete installation can cost \$250,000, which still presents a significant barrier to access.

\begin{figure}[t]
    \centering
    \includegraphics[width=0.48\textwidth]{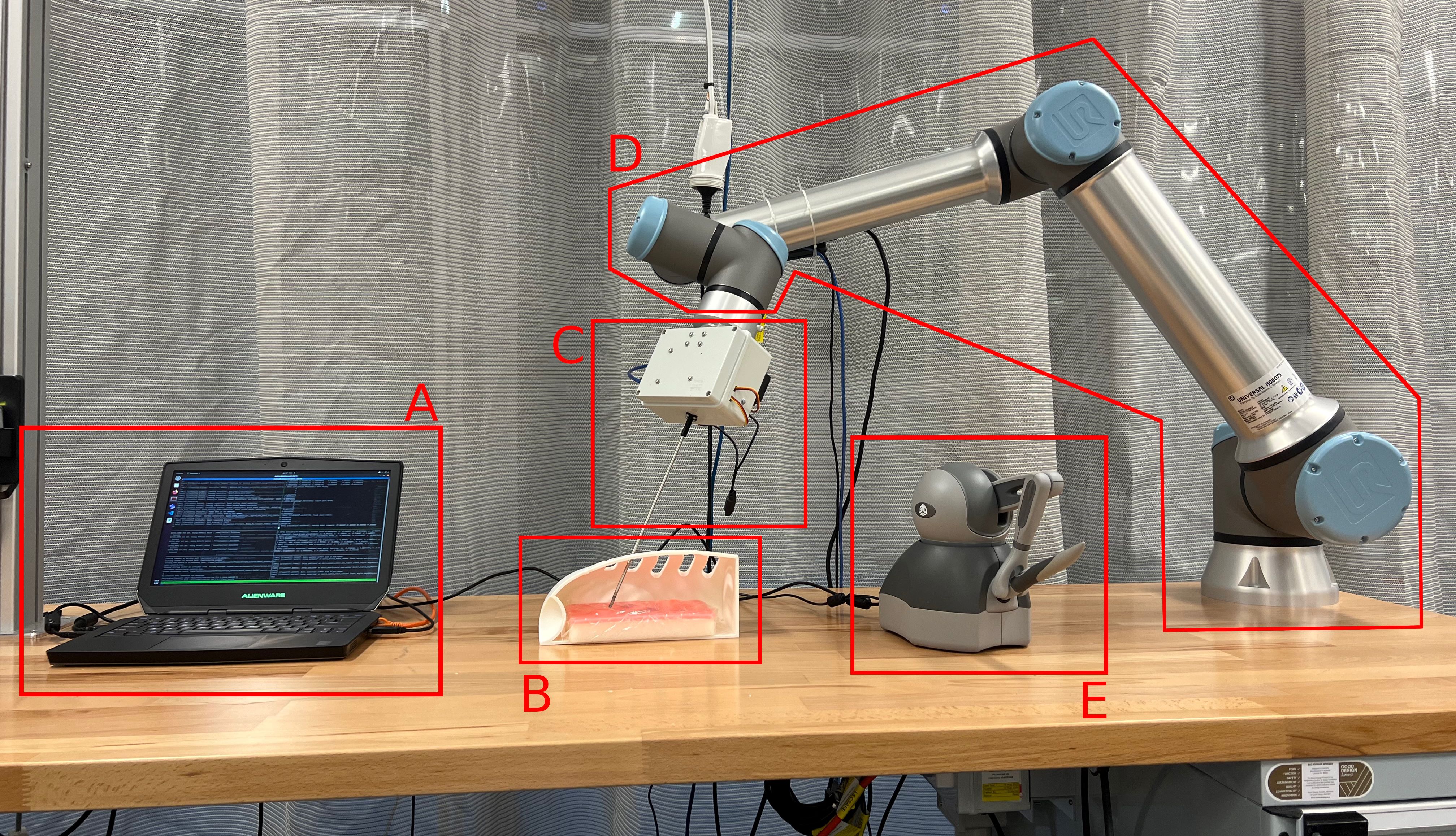}
    \caption{The assembled macro-micro system during the phantom trial, with (\textbf{A}) the control computer, (\textbf{B}) the phantom chest model, (\textbf{C}) the micro manipulator, (\textbf{D}) the UR5e arm and (\textbf{E}) the Touch stylus device.}
    \label{fig:MacroMicroAssembly}
\end{figure}

This paper presents a low-cost prototype micro-scale manipulator module (micro module) based on the SnakeRaven surgical manipulator designed by Razjigaev \textit{et al.} \cite{snakeraven}. The custom manipulator is designed to be attached to the Universal Robots UR5e 6-axis robotic arm, which acts as the macro-scale manipulator (macro module). The conjoined manipulators can be teleoperated in tandem using a 3D Systems Touch haptic stylus device. The complete system is depicted in Fig.~\ref{fig:MacroMicroAssembly}. Communication with the stylus and stable control of the robotic system is achieved using custom software which has been developed using open-source or freely available device driver packages. It is hoped that this work will serve as an accessible proof-of-concept and information resource for further efforts in the field of robotic minimally invasive surgery.

The source code, CAD files, a URDF model of the micro module, and a complete parts list can be found at: \href{https://github.com/drliaowu/MacroMicroSurgicalRobot}{https://github.com/drliaowu/MacroMicroSurgicalRobot}, and a demonstration video can be found at: \href{https://youtu.be/NCoqgnThfMY}{https://youtu.be/NCoqgnThfMY}. 
The estimated total cost of the micro manipulator assembly is \$252.28 AUD.

\section{Micro Module Design}
\begin{figure}[th!]
    \centering
    \includegraphics[width=0.4\textwidth]{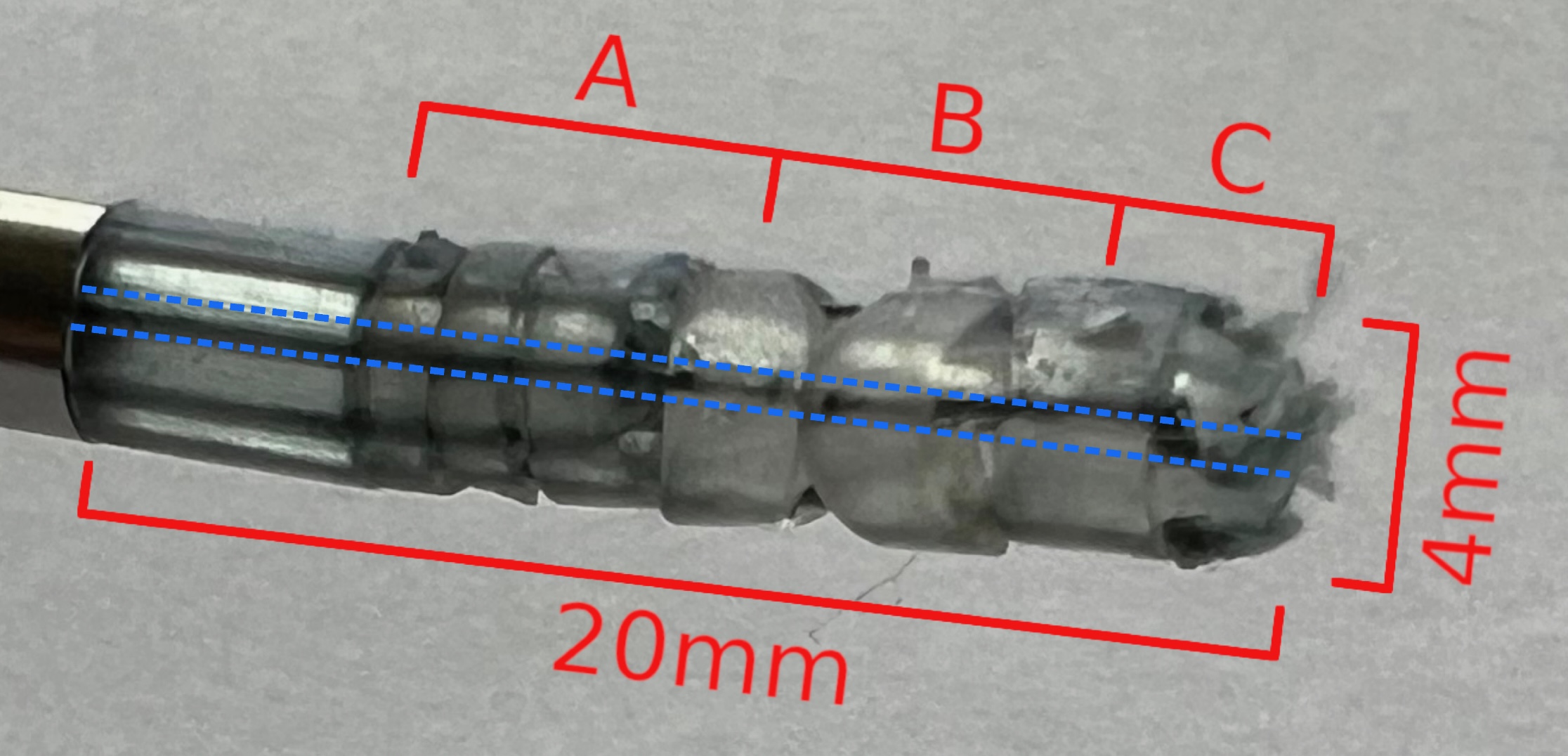}
    \caption{A close-up image of the tip of the manipulator showing the proximal rolling joints (\textbf{A}), distal rolling joints (\textbf{B}), and end-effector (\textbf{C}). The path of the central lumen is outlined in blue.}
    \label{fig:EndEffectorCloseUp}
\end{figure}

\begin{figure}[th!]
    \centering
    \includegraphics[width=0.5\textwidth]{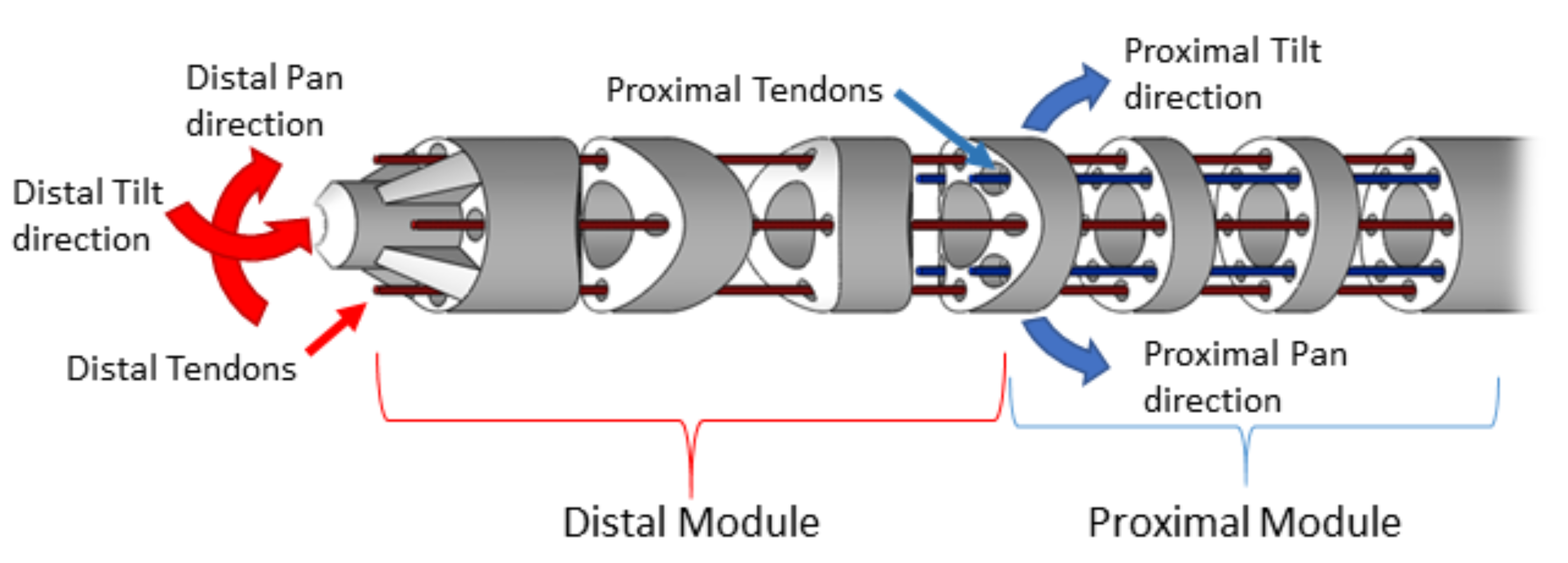}
    \caption{A diagram of the manipulator showing the tendon routing paths, and the pan and tilt directions for the proximal and distal modules.\cite{snakeraven}}
    \label{fig:ManipulatorDiagram}
\end{figure}

The end-effector of the micro module can be seen in fig.~\ref{fig:EndEffectorCloseUp}, and its structure and motion are detailed in fig.~\ref{fig:ManipulatorDiagram}. It consists of a 20mm long, 4mm diameter, tendon-driven continuum manipulator which is fixed to a 250mm long steel tube of the same diameter. A 2mm central lumen passes through each rolling joint unit and is designed to accommodate a compact surgical instrument or sensor, such as an endoscope.

\begin{figure}[th!]
    \centering
    \includegraphics[width=0.4\textwidth]{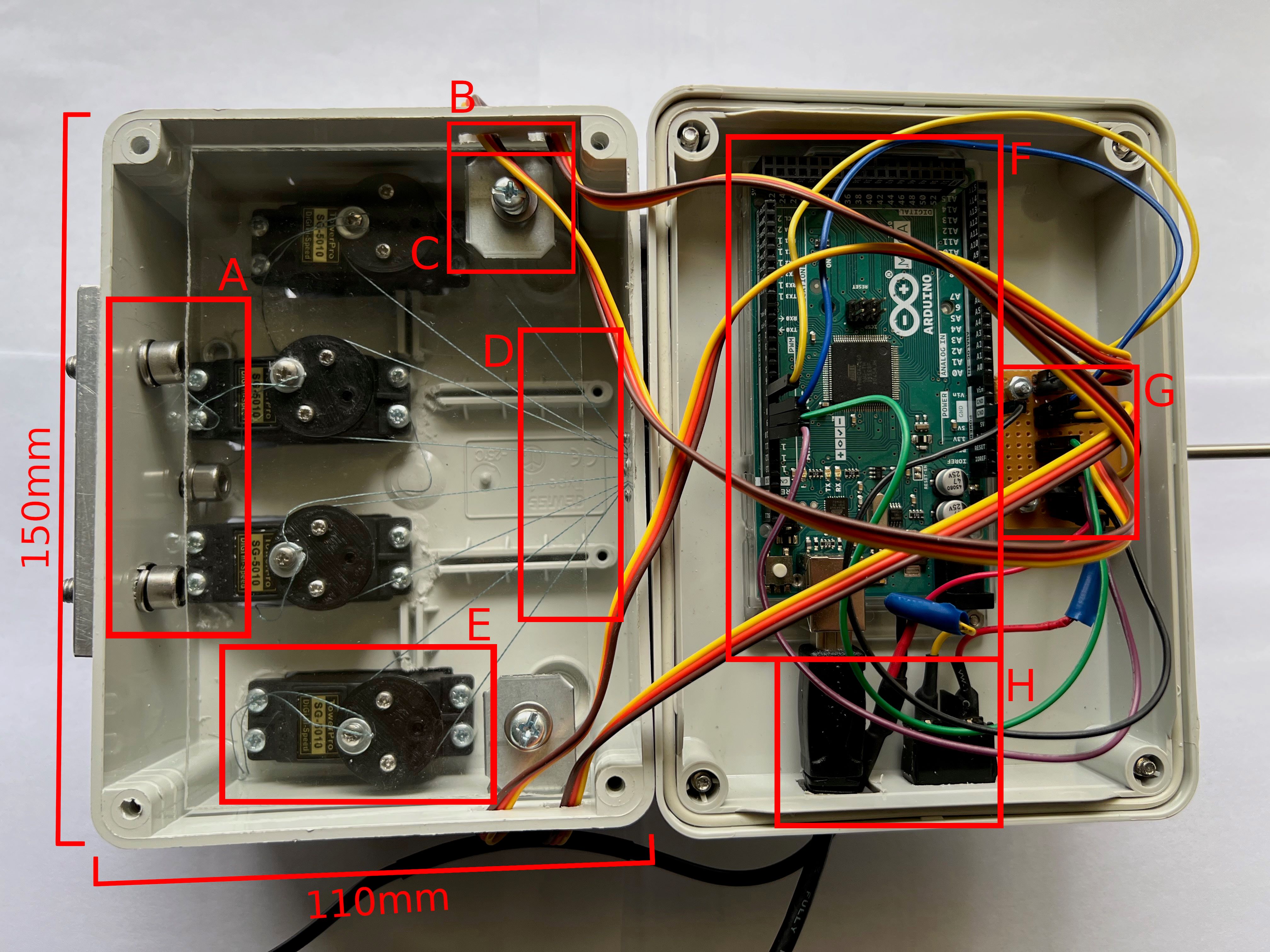}
    \caption{A top view of the interior of the micro module showing the UR5e mounting point (\textbf{A}), servo cable channels (\textbf{B}), servo shield supports (\textbf{C}), tendons with instrument channel entry point at right (\textbf{D}), a servo and pulley assembly (\textbf{E}), the Arduino Mega control board (\textbf{F}), the servo interface board (\textbf{G}), and power and data cable entry with power switch (\textbf{H}).}
    \label{fig:MicroModuleOpenTop}
\end{figure}

A rectangular enclosure at the distal end of the tube, shown in fig.~\ref{fig:MicroModuleOpenTop}, contains the driving pulleys and actuators, the Arduino Mega control board, a servo control and power interface board, as well as any instrumentation such as the endoscopic camera used during demonstration procedures. The manipulator consists of five discrete rolling-joint nodes, each with a rolling surface on two opposing faces. The nodes form two sub-modules - proximal and distal - distinguished by the radius of curvature of their rolling surfaces. Each sub-module provides 2 degrees of freedom (DOF), and is offset from the other by 45 degrees along the z-axis. The design parameters of the manipulator are shown in table \ref{tab:ManipulatorParameters}. These parameters are represented using the convention established by Razjigaev \textit{et al}. wherein a given module contains $n$ discrete rolling joints of width (diameter) $w$, and $\alpha$ and $d$ represent the half-angle of curvature of, and the distance between, rolling surfaces on each joint respectively. The values mirror those used for the SnakeRaven manipulator since these have previously been validated for surgical applications \cite{snakeraven,Razjigaev2022}.

The manipulator's rolling joints, and accompanying parts including the end-effector and the base adaptor between the steel tube and the first rolling surface, were fabricated using a FormLabs Form 3 stereolithography (SLA) 3D printer. The parts, shown in Fig.~\ref{fig:EndEffectorCloseUp} were printed in FormLabs clear photopolymer at a resolution of 25$\mu$m. The photopolymer exhibits sufficient strength for use in a prototype surgical manipulator, with an ultimate tensile strength of 65MPa and a Notched Izod impact strength of 25J/m after being UV-cured\cite{FormLabsResinInfo}.
Clear resin was also chosen to increase the ease with which driving tendons could be threaded through the 0.5mm channels in each part during assembly and to allow tendons to be monitored for signs of wear during testing.

\begin{table}[t]
    \centering
    \caption{The design parameters of the micro-module manipulator.}
        \begin{tabular}{c|c|c|c|c }
            \hline
            Module & n & $w$ (mm) & $\alpha$ (rad) & $d$ (mm)\\ \hline
            Proximal & 3 & 4 & 0.2 & 1\\ \hline
            Distal & 3 & 4 & 0.88 & 1\\ \hline
        \end{tabular}
    \label{tab:ManipulatorParameters}
\end{table}

The pulley wheels of the manipulator, shown in Fig.~\ref{fig:MicroModuleOpenTop} were manufactured using the Fused Deposition Modelling (FDM) printing process on a Creality Ender 3 printer. 
The pulleys feature a pair of channels that allow the attached tendons to be threaded through and anchored in place using a standard M3 screw and washer set. This simple mechanism facilitates the adjustment of tendon tension after assembly while fixing tendons in place during operation.

A mounting bracket consisting of a 5mm thick machined aluminum plate with three countersunk bolt holes is fixed to the back face of the manipulator, enabling it to be attached to the tool flange of the UR5e arm.

\section{Robotic System Architecture}
\subsection{Hardware}
\begin{figure}[t]
    \centering
    \includegraphics[width=0.4\textwidth]{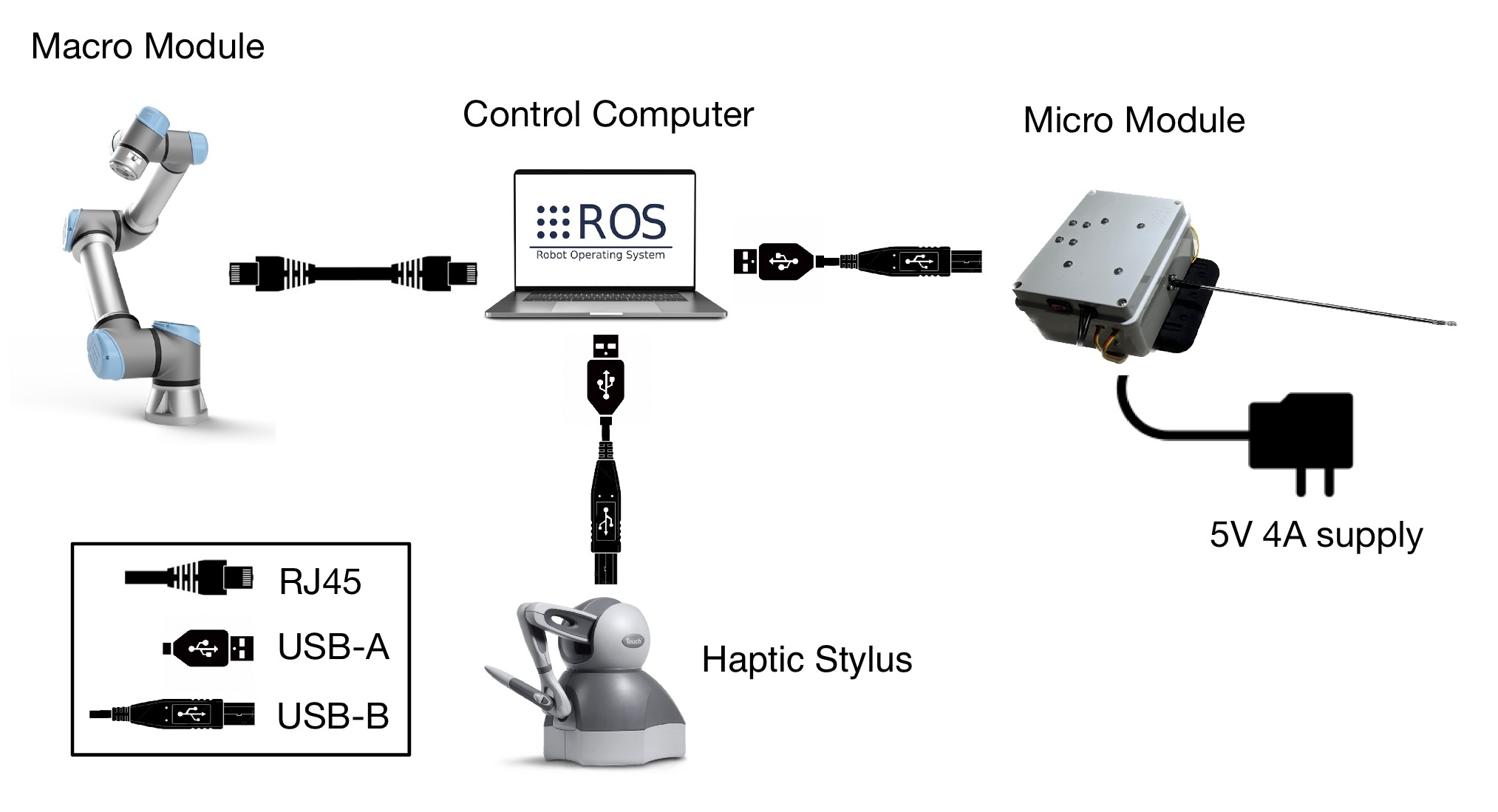}
    \caption{The system hardware components and their interfaces.}
    \label{fig:HardwareInterfaces}
\end{figure}

The complete macro-micro robotic system consists of the custom micro-manipulator, the UR5e robotic arm, and the 3DSystems Touch haptic stylus. Each is connected to a central computer running ROS (the Robotic Operating System). The control computer is equipped with an Intel i5 4210U CPU (4 cores, 1.7GHz) and 16GB of RAM. The major system components and their connections are depicted in Fig.~\ref{fig:HardwareInterfaces}.

\subsection{Software}
\begin{figure}[t]
    \centering
    \includegraphics[width=0.4\textwidth]{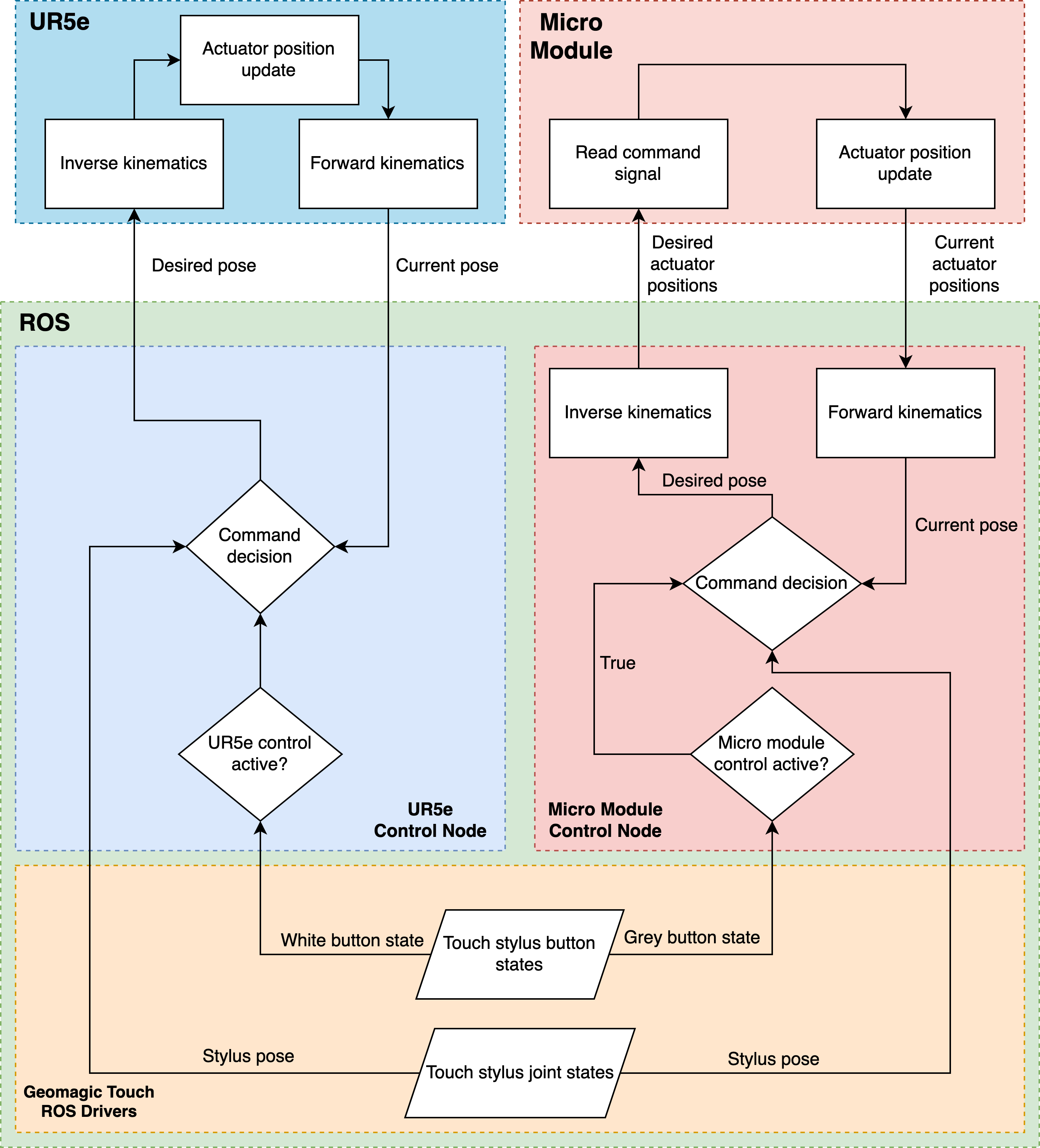}
    \caption{The architecture of the control software for the complete macro-micro system.}
    \label{fig:FinalSystemArchitecture}
\end{figure}

Fig.~\ref{fig:FinalSystemArchitecture} depicts the structure and logical flow of the system's ROS control software, which was written in C++. The pose and button states of the stylus are provided to the UR5e and Micro Module control systems. The white button is assigned to control the activation state of the macro module's teleoperation, while the grey button performs the same function for the micro module. 
Teleoperation of the modules can be activated simultaneously or independently, although, in an operational context, the macro manipulator would likely be positioned first and remain largely stationary as a tool affixed to the micro manipulator is used to perform the MIS procedure in-situ.
Each module's control node calculates the desired pose of its respective manipulator. Forward and inverse kinematic calculations for the UR5e are performed by its dedicated control computer. Those for the micro module are performed by the corresponding ROS node.

\subsection{Haptic Stylus Interface}
The interface between the primary control system and the 3DSystems Touch haptic stylus was achieved via a dedicated ROS node. 
The node utilizes the Geomagic Touch ROS Drivers package developed by Mathur \textit{et al.} \cite{MathurTeleop,MathurControlStrategies}. 
The real-time pose and translational velocity of the stylus' joints are extracted, and kinematic calculations are performed in order to enable teleoperated control of the macro and micro modules.

\subsection{Macro Module Interface}
Teleoperation of the macro module, the Universal Robots UR5e robotic arm, was achieved using another ROS node. Stylus pose data is obtained by subscribing to the Touch driver node, which can then be compared with the current state of the arm to determine its
trajectory in upcoming time frames.

Upon the activation of teleoperation via the dedicated control button on the Touch stylus, the difference in position and
orientation of the stylus compared to its initial pose is tracked in real time. This pose difference is
scaled and applied to the pose of the macro module, yielding a new pose that causes the arm's end-effector to mirror the trajectory
of the stylus over time.

\subsection{Micro Module Interface}
Control of the micro module was achieved by combining a ROS node with an Arduino control program receiving angular position commands via a TCP connection. The theoretical framework and MATLAB programs developed during the SnakeRaven project \cite{snakeraven} served as a reference for the control node. Given the project's limited time-frame and that its primary objective was to demonstrate the teleoperation of a macro-micro system, it was deemed sufficient to achieve predictable, intuitive motion of the micro-manipulator at the expense of accuracy. Further work to refine each sub-system should address the reliability and accuracy of the manipulator's motion in response to stylus input.

The ROS control node employs the Damped Least Squares method for inversion of the Jacobian in order to translate manipulator pose updates from the joint space to the tendon space. This inverse kinematic method possesses the advantage of enabling the inversion of non-square matrices while dampening the magnitude of motions near singularities and joint angle limits, enhancing safety\cite{ImprovedDampedLeastSquares,BussIntroToIK,ChiaveriniLeastSquares,snakeraven}.

\begin{figure}[th!]
    \centering
    \includegraphics[width=0.3\textwidth]{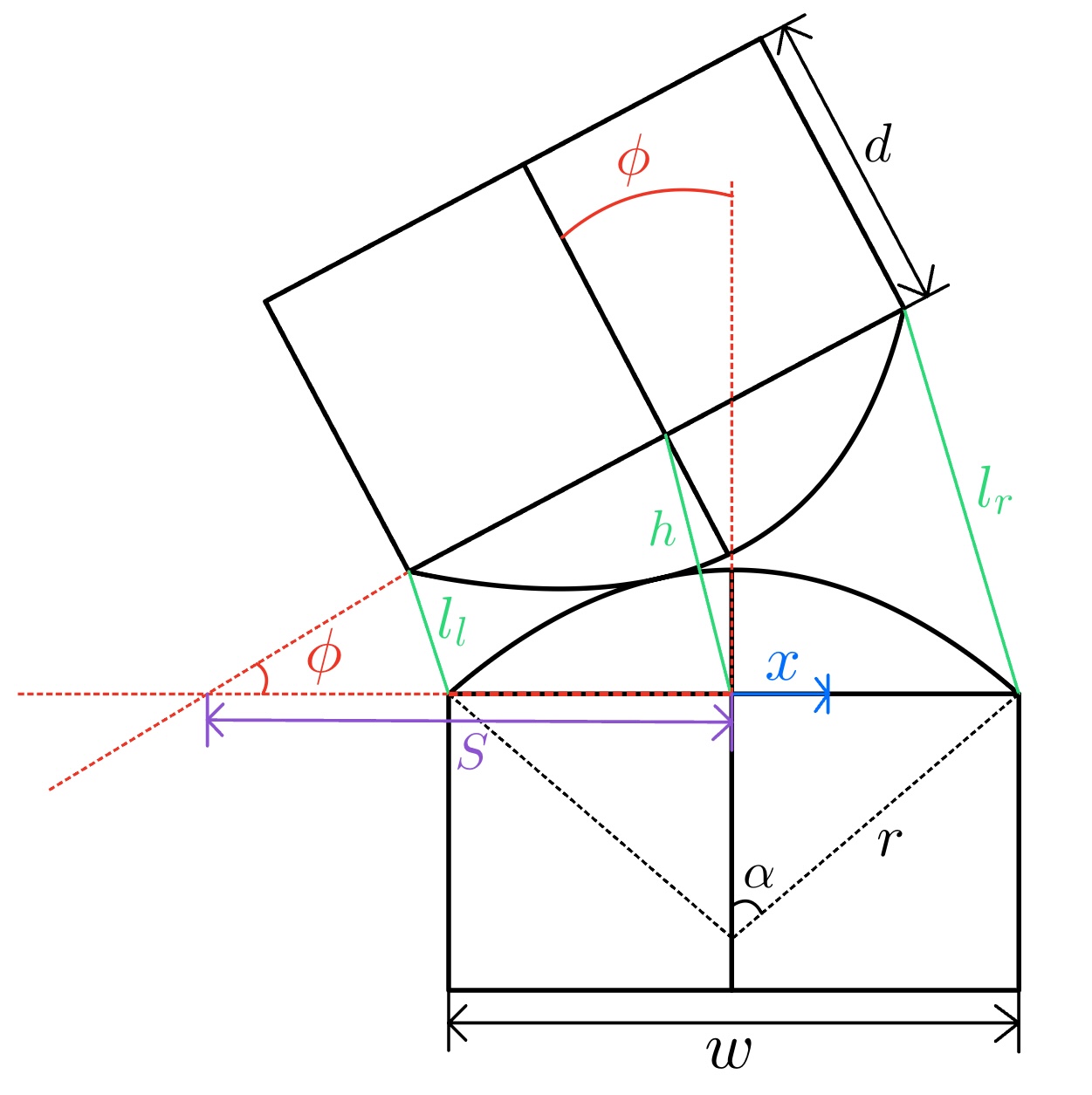}
    \caption{A diagram of a rolling joint interface with parameters half-angle of curvature $\alpha$, width $w$ and separation distance $d$}.
    \label{fig:JointDiagram}
\end{figure}

The ``Actuator Jacobian" method outlined by Razjigaev \textit{et al}. \cite{snakeraven} was used to translate from the tendon space to the actuator space. Fig. \ref{fig:JointDiagram} relates the design parameters $\alpha$, $w$ and $d$ at a joint interface to the key control parameters: joint angle $\phi$, as well as $l_l$ and $l_r$, the left and right tendon lengths respectively. $x$, $h$, $r$, and $S$ are intermediate variables used in deriving the key equation \ref{eqn:deltaL}, which relates $\phi$ to the corresponding change in left tendon length $\Delta{l_l}$ for a given joint.

\begin{equation}
    \begin{split}
        \Delta l&=l(\phi)-l(0)\\
        &=2r(\cos(\alpha)-\cos(\alpha-\frac{\phi}{2}))
    \end{split}
    \label{eqn:deltaL}
\end{equation}

Control of the manipulator's pose is achieved by iteratively modulating the combined tendon lengths via pulleys according to the desired change at each joint. Parameters of the physical system, as well as control gain and scaling factors, are exposed and readily modifiable in the control node to suit varied applications.

\section{Results and Discussion}
The performance of the completed robotic system was evaluated both on a per-module basis and as an integrated whole. 
Isolated trials were conducted on the macro and micro-modules to assess the accuracy and intuitiveness of teleoperated control for each, allowing preliminary calibration or tuning issues to be identified and eliminated. 
Once the performance of the isolated systems was deemed acceptable, they could be teleoperated in conjunction to demonstrate their utility in maneuvering an endoscopic camera fitted to the end-effector.

\subsection{Macro Module Teleoperation}

\begin{figure}[th!]
    \centering
    \includegraphics[width=0.46\textwidth]{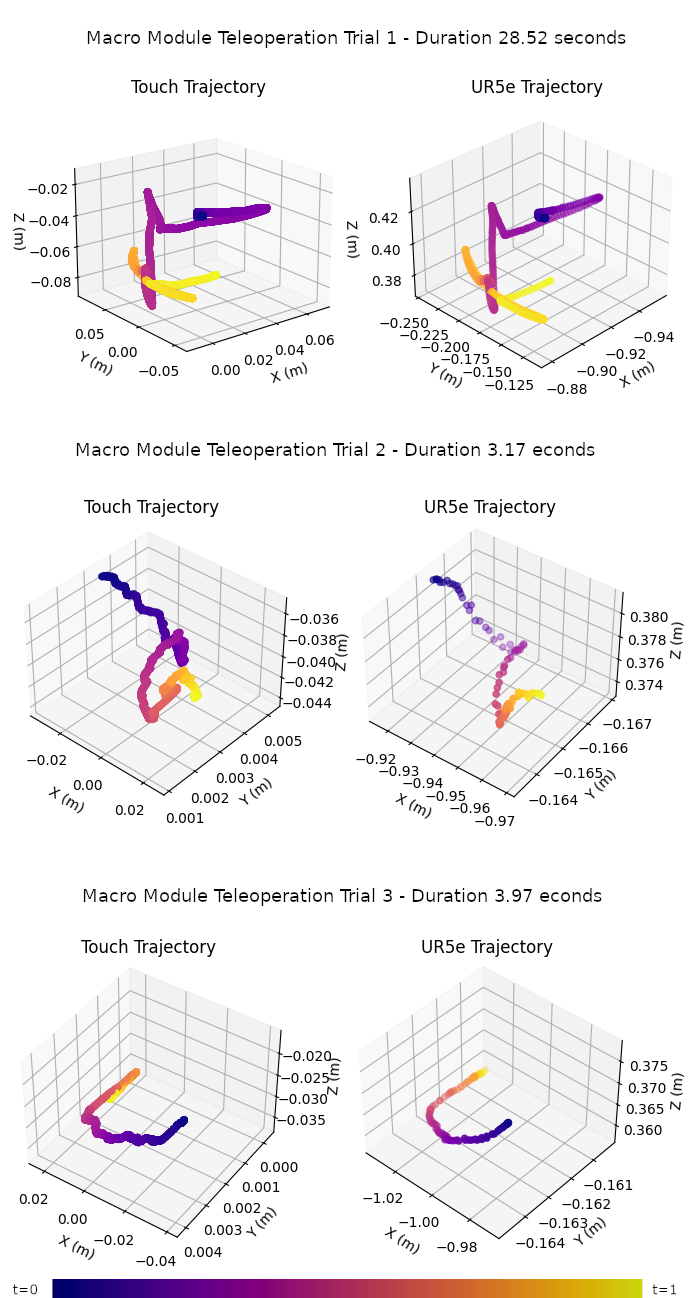}
    \caption{The trajectories of the Touch stylus (left) and the UR5e end-effector (right) in the first, second, and third teleoperation trials.}
    \label{fig:MacroTrialsCombined}
\end{figure}

Fig.~\ref{fig:MacroTrialsCombined} depicts the results of the UR5e teleoperation trials. 
Each graph set represents a trajectory defined by the manual movement of the Touch stylus, which was then mirrored by the UR5e. 
The progression of time along the trajectories in each trial is indicated by the color gradient of the path. 
The key at the base of the figure indicates how this should be interpreted, where t=0 and t=1 are the start and end of each trajectory, respectively. 
The discrepancy in point density between the Touch and UR5e paths is attributed to the differing communication polling rates of the two systems. 
The state of the Touch Stylus was published at a rate of 1kHz, while the state of the UR5e was published at 100Hz.

As is evident in each plot, the UR5e successfully mirrored the trajectory of the stylus. 
During the operation, it was noted that the arm exhibited marked vibrations during translational motion. 
However, these vibrations do not appear to have significantly affected the overall similarity of the arm’s trajectory with that of the Touch. 
The orientation of the arm also precisely followed that of the Touch stylus during the experiment, although further, targeted testing and analysis of the orientation control is warranted to establish its accuracy.

\subsection{Micro Module Teleoperation}

\begin{figure}[th!]
    \centering
    \includegraphics[width=0.3\textwidth]{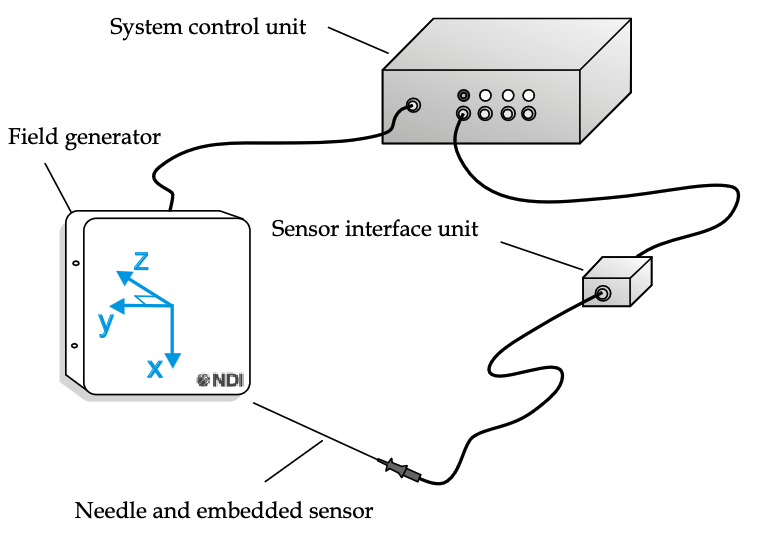}
    \caption{A diagram of the Aurora electromagnetic tracking system, depicting the tracking needle and the coordinate
    axes of the field generator \cite{BoutalebEMTracking}.}
    \label{fig:AuroraDiagram}
\end{figure}

The effectiveness of the micro module teleoperation system was assessed in a series of trials using an NDI Aurora electromagnetic tracker since the manipulator's end effector lacks an intrinsic method of direct motion tracking. 
The tracker and its peripheral control and communication hardware are depicted in Fig.~\ref{fig:AuroraDiagram}. 

The tracker measured the displacement of a sensor needle fixed to the micro manipulator's end effector while it was guided through varying trajectories using the Touch stylus.

\begin{figure}[th!]
    \centering
    \includegraphics[width=0.46\textwidth]{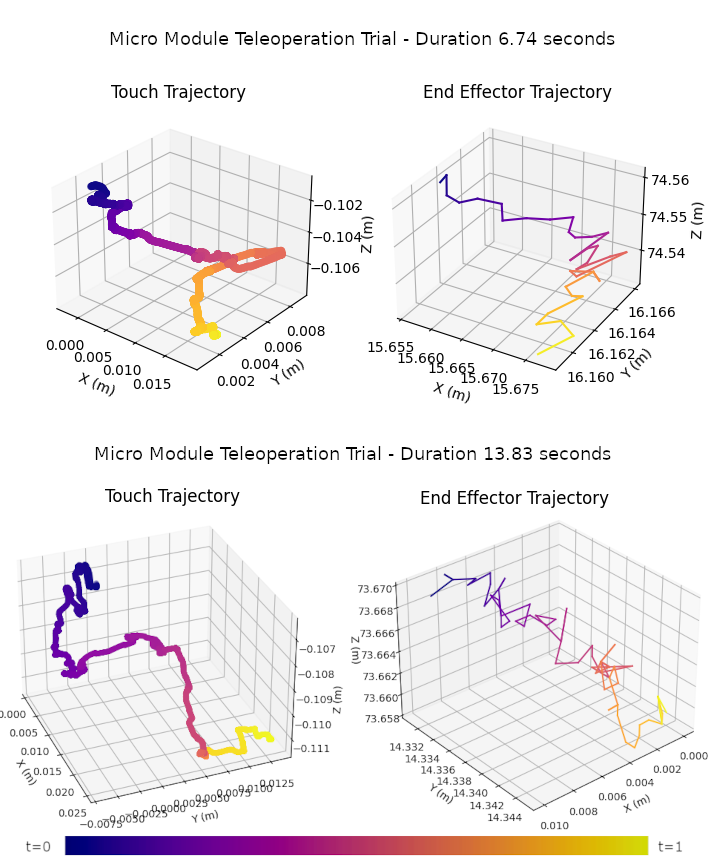}
    \caption{The trajectories of the Touch stylus (left) and the Micro Module end-effector (right) in the first and second teleoperation trials.}
    \label{fig:MicroTrialsCombined}
\end{figure}

The results of the trials are depicted in Fig.~\ref{fig:MicroTrialsCombined}. 
The average absolute error in position tracking reported by the Aurora during the first group of trials was 0.0272mm and 0.0161mm in the second. 

Similar to the experiments conducted with the macro module,
the disparity in polling rates between the Touch (1kHz) and the Aurora (40Hz) resulted in a notable distinction between
the resolutions of the final plots. Another factor influencing the quality of
measurements was the tendency of the Aurora to report a ``Bad Fit'' for a brief period of time. An attempt to mitigate
this occurrence was made, by positioning the manipulator tip approximately 10cm from the field generator's surface and
with the tracking needle's axis perpendicular to the generator's face. However, as a combined result of these
compromising factors, a large number of plots generated in the first set of trials did not possess a sufficient number
of data points to give accurate representations of the end effector's trajectory. Those plots with the most distinct
manipulator trajectories have been selected for discussion.

During the trials, it was noted that the micro module occasionally exhibits a significant degree of latency between
the incidence of a commanded trajectory via the stylus and the response of the manipulator in mirroring the pose. This issue can
likely be addressed by refining the manipulator parameters of the micro module's ROS control node. 
Precise measurements of the true curvature angle and tendon separation of each module will reduce errors due to inaccuracy in the physical models. 
Additionally, the Arduino Mega may be replaced with a control board designed specifically for high-precision real-time applications.

\subsection{Combined Teleoperation}
Once the teleoperation of the individual macro and micro modules had been developed and validated, the two systems could be combined and teleoperated as a single unit. 
The performance of the macro-micro system was validated via a demonstration, in the form of a simulated tumour biopsy using an anatomical phantom chest. The experimental setup is depicted in figure \ref{fig:MacroMicroAssembly}.
The micro module was mounted to the tool flange of the UR5e, and an endoscopic camera was installed through the central lumen of the end-effector to enable imaging of the tumour site. The trial operation involved starting from a horizontal position above the phantom, guiding the end-effector to navigate between the ribs of the phantom model, acquiring footage of the tumour site, then retracting the end-effector through the same opening.

Teleoperation of the combined macro-micro system, with 10 DOF, proved successful. The average time taken to complete the trial operation was 5 minutes and 34 seconds. Each module could be controlled independently via the Touch stylus. 
The macro module provided a large effective workspace within which to position the endoscopic instrument. The arm offered free motion along all six degrees of freedom, while providing a stable base from which the position of the micro module could be finely adjusted.

Similar to the macro module, the micro module could be freely positioned but was rigidly fixed in place when teleoperation was deactivated, and its positioning was robust to small manually applied external forces.
Control latency and motion scaling factors in both modules was observed to be adequate to allow for intuitive positioning, and navigation through a narrow opening between the ribs of the phantom model.
However, further tuning of control system parameters for both the UR5e and the micro manipulator is expected to significantly improve teleoperation performance. Footage of one of the phantom trials is available in the demonstration video.

\section{Conclusion}
An additively manufactured, tendon-driven continuum manipulator modeled after the SnakeRaven manipulator design\cite{snakeraven,Razjigaev2022} has been constructed. 
The manipulator has been shown to be capable of effective teleoperative control via a 3DSystems Touch haptic stylus device. 
Its performance has been validated using an electromagnetic tracking device.
It has been successfully teleoperated as part of a macro-micro system formed by attaching it to a commercially available robotic arm, which was then validated by a simulated biopsy procedure.

Building upon open-source and/or freely-available software packages, the control software is designed to facilitate rapid prototyping of this or similar systems. 
It is hoped that the low-cost and modular nature of the hardware and software comprising the system will contribute to addressing the prohibitive cost, low availability, and proprietary nature of most surgical robotic platforms described in the literature, and encourage subsequent research in the areas of macro-micro robotic systems, snake-like manipulators, and teleoperated control.

\section{Future Work}
Significant improvements and additions to the manipulator system can be realised with future research and development efforts. Subsequent work may be conducted to address the issue of vibrations observed in the
translational motion of the arm in response to teleoperation commands. While these vibrations did not significantly
affect the pose-mirroring performance illustrated in Fig.~\ref{fig:MacroTrialsCombined} and Fig.~\ref{fig:MicroTrialsCombined}, they should be eliminated by further tuning of control parameters to ensure safe and reliable maneuverability in an operating environment.
Additional research attention could be focused on developing a modular end-effector system for the micro-module. 
Presently, the removal and replacement of the tendon-driven end-effector is a complicated process. 
Increasing the ease with which various continuum end-effector designs can be installed and removed from the actuator base would serve to increase its suitability for patient-specific surgical interventions.

Further work can be conducted on the integrated system to provide haptic feedback to the operator via
the Touch Stylus. Such feedback could serve to alert users that they have reached the joint
limits of the manipulator, providing the operator with an awareness of the workspace boundaries of the surgical equipment.
With additional sensor integration, haptic feedback could also be employed to convey the
resistance forces encountered by the end-effector. This feedback would ensure
safe and intuitive operation while still conferring the improved precision and remote operation capabilities of robotic surgical devices.

\addtolength{\textheight}{-12cm}   

\section*{ACKNOWLEDGMENT}
Thank you to Stephen Kuhle for assisting in the construction of the micro module. Thanks also to Samuel Hall and Mustafa Sevinc for providing advice in interfacing hardware with ROS. Thank you to Miranda Yuan for her invaluable support and advice.

\bibliographystyle{IEEEtran}
\bibliography{IEEEabrv,References.bib}

\begin{thebibliography}{10}
\providecommand{\url}[1]{#1}
\csname url@rmstyle\endcsname
\providecommand{\newblock}{\relax}
\providecommand{\bibinfo}[2]{#2}
\providecommand\BIBentrySTDinterwordspacing{\spaceskip=0pt\relax}
\providecommand\BIBentryALTinterwordstretchfactor{4}
\providecommand\BIBentryALTinterwordspacing{\spaceskip=\fontdimen2\font plus
\BIBentryALTinterwordstretchfactor\fontdimen3\font minus
  \fontdimen4\font\relax}
\providecommand\BIBforeignlanguage[2]{{%
\expandafter\ifx\csname l@#1\endcsname\relax
\typeout{** WARNING: IEEEtran.bst: No hyphenation pattern has been}%
\typeout{** loaded for the language `#1'. Using the pattern for}%
\typeout{** the default language instead.}%
\else
\language=\csname l@#1\endcsname
\fi
#2}}

\bibitem{ClarkOpenSource}
A.~B. Clark, V.~Mathivannan, and N.~Rojas, ``A continuum manipulator for
  open-source surgical robotics research and shared development,'' \emph{IEEE
  Transactions on Medical Robotics and Bionics}, vol.~3, no.~1, pp. 277--280,
  2021.

\bibitem{surgicalrobotsandcomputer}
R.~H. Taylor, N.~Simaan, A.~Menciassi, and G.-Z. Yang, ``Surgical robotics and
  computer-integrated interventional medicine [scanning the issue],''
  \emph{Proceedings of the IEEE}, vol. 110, no.~7, p. 823 – 834, 2022.

\bibitem{yangmacromicro}
T.~W. Yang, W.~L. Xu, and J.~D. Han, ``Dynamic compensation control of flexible
  macro–micro manipulator systems,'' \emph{IEEE Transactions on Control
  Systems Technology}, vol.~18, no.~1, pp. 143--151, 2010.

\bibitem{continuumrobotssurvey}
J.~Burgner-Kahrs, D.~C. Rucker, and H.~Choset, ``Continuum robots for medical
  applications: A survey,'' \emph{IEEE Transactions on Robotics}, vol.~31,
  no.~6, pp. 1261--1280, 2015.

\bibitem{yousefmacromicro}
B.~Yousef, R.~Patel, and M.~Moallem, ``A macro-robot manipulator for medical
  applications,'' vol.~1, 2006, Conference paper, p. 530 – 535.

\bibitem{Telesurgery}
S.-B. Xia and Q.-S. Lu, ``Development status of telesurgery robotic system,''
  \emph{Chinese Journal of Traumatology}, vol.~24, no.~3, pp. 144--147, 2021.

\bibitem{LiuHaptics}
T.~Liu, T.~Zhang, J.~Katupitiya, J.~Wang, and L.~Wu, ``Haptics-enabled forceps
  with multimodal force sensing: Toward task-autonomous surgery,''
  \emph{IEEE/ASME Transactions on Mechatronics}, pp. 1--12, 2023.

\bibitem{OptimalVision}
A.~Razjigaev, A.~Pandey, D.~Howard, J.~Roberts, A.~Jaiprakash, R.~Crawford, and
  L.~Wu, ``Optimal vision-based orientation steering control for a 3-d printed
  dexterous snake-like manipulator to assist teleoperation,'' \emph{IEEE/ASME
  Transactions on Mechatronics}, pp. 1--12, 2023.

\bibitem{ChenOpenSource}
Z.~Chen, A.~Deguet, R.~Taylor, S.~DiMaio, G.~Fischer, and P.~Kazanzides, ``An
  open-source hardware and software platform for telesurgical robotics
  research,'' \emph{The MIDAS Journal}, 08 2013.

\bibitem{raven}
B.~Hannaford, J.~Rosen, D.~W. Friedman, H.~King, P.~Roan, L.~Cheng, D.~Glozman,
  J.~Ma, S.~N. Kosari, and L.~White, ``Raven-ii: An open platform for surgical
  robotics research,'' \emph{IEEE Transactions on Biomedical Engineering},
  vol.~60, no.~4, pp. 954--959, 2013.

\bibitem{snakeraven}
A.~Razjigaev, A.~K. Pandey, D.~Howard, J.~Roberts, and L.~Wu, ``Snakeraven:
  Teleoperation of a 3d printed snake-like manipulator integrated to the raven
  ii surgical robot,'' in \emph{2021 IEEE/RSJ International Conference on
  Intelligent Robots and Systems (IROS)}.\hskip 1em plus 0.5em minus
  0.4em\relax IEEE, 2021, pp. 5282--5288.

\bibitem{Razjigaev2022}
------, ``End-to-end design of bespoke, dexterous snake-like surgical robots: A
  case study with the raven ii,'' \emph{IEEE Transactions on Robotics},
  vol.~38, no.~5, pp. 2827--2840, 2022.

\bibitem{FormLabsResinInfo}
\BIBentryALTinterwordspacing
``General purpose resins - materials for high resolution models and rapid
  prototyping.'' [Online]. Available:
  \url{https://formlabs-media.formlabs.com/datasheets/1801089-TDS-ENUS-0P.pdf}
\BIBentrySTDinterwordspacing

\bibitem{MathurTeleop}
B.~Mathur, A.~Topiwala, S.~Schaffer, M.~Kam, H.~Saeidi, T.~Fleiter, and
  A.~Krieger, ``A semi-autonomous robotic system for remote trauma
  assessment,'' in \emph{2019 IEEE 19th International Conference on
  Bioinformatics and Bioengineering (BIBE)}, 2019, pp. 649--656.

\bibitem{MathurControlStrategies}
\BIBentryALTinterwordspacing
B.~Mathur, A.~Topiwala, H.~Saeidi, T.~Fleiter, and A.~Krieger, \emph{Evaluation
  of Control Strategies for a Tele-manipulated Robotic System for Remote Trauma
  Assessment}, pp. 7--14. [Online]. Available:
  \url{https://epubs.siam.org/doi/abs/10.1137/1.9781611975758.2}
\BIBentrySTDinterwordspacing

\bibitem{ImprovedDampedLeastSquares}
M.~Na, B.~Yang, and P.~Jia, ``Improved damped least squares solution with joint
  limits, joint weights and comfortable criteria for controlling human-like
  figures,'' in \emph{2008 IEEE Conference on Robotics, Automation and
  Mechatronics}, 2008, pp. 1090--1095.

\bibitem{BussIntroToIK}
S.~Buss, ``Introduction to inverse kinematics with jacobian transpose,
  pseudoinverse and damped least squares methods,'' \emph{IEEE Transactions in
  Robotics and Automation}, vol.~17, 05 2004.

\bibitem{ChiaveriniLeastSquares}
S.~Chiaverini, B.~Siciliano, and O.~Egeland, ``Review of the damped
  least-squares inverse kinematics with experiments on an industrial robot
  manipulator,'' \emph{IEEE Transactions on Control Systems Technology},
  vol.~2, no.~2, pp. 123--134, 1994.

\bibitem{BoutalebEMTracking}
S.~Boutaleb, E.~Racine, O.~Fillion, A.~Bonillas, G.~Hautvast, D.~Binnekamp, and
  L.~Beaulieu, ``Performance and suitability assessment of a real-time 3d
  electromagnetic needle tracking system for interstitial brachytherapy,''
  \emph{Journal of Contemporary Brachytherapy}, vol.~7, no.~4, p. 280 – 289,
  2015.

\end{thebibliography}

\end{document}